\pdfoutput=1

\documentclass[11pt]{article}

\usepackage{acl}
\usepackage{nert}

\usepackage{tikz}

\usepackage{times}
\usepackage{latexsym}

\usepackage{algorithm}
\usepackage{algorithmic}

\usepackage[T1]{fontenc}

\usepackage{microtype}

\newcommand{\amr}[1]{\texttt{#1}}

\newcommand{\repname}{\textsc{DocAMR}\xspace}
\newcommand{\evlname}{\textsc{DocSmatch}\xspace}
\title{\repname: Multi-Sentence AMR Representation and Evaluation}

\newcommand{\neghphantom}[1]{\settowidth{\dimen0}{#1}\hspace*{-.5\dimen0}}
\newcommand{\affilsym}[1]{$^{#1}$\hspace{1pt}\neghphantom{$^{#1}$}} 

\author{Tahira Naseem\affilsym{\clubsuit,*}\quad 
Austin Blodgett\affilsym{\spadesuit}\quad 
Sadhana Kumaravel\affilsym{\clubsuit}
\\
  \textbf{Tim O'Gorman\affilsym{\heartsuit}\quad Young-Suk Lee\affilsym{\clubsuit}\quad Jeffrey Flanigan\affilsym{\diamondsuit}\quad Ramón Fernandez Astudillo\affilsym{\clubsuit}} 
  \\  
  \textbf{Radu Florian\affilsym{\clubsuit}\quad  Salim Roukos\affilsym{\clubsuit}\quad
  Nathan Schneider$^{\triangle}$ } \\ $^{*}$\eml{tnaseem@us.ibm.com} \\
  $^{\clubsuit}$IBM Research AI\quad
  $^{\spadesuit}$U.S. Army Research Laboratory, Adelphi, MD \\ 
  $^{\heartsuit}$Thorn\quad
  $^{\diamondsuit}$University of California, Santa Cruz\quad
  $^{\triangle}$Georgetown University
  }

\begin{document}
\maketitle

\begin{abstract}

Despite extensive research on parsing of English \emph{sentences} into Abstraction Meaning Representation (AMR) graphs, 
\emph{full-document} parsing into a unified graph representation lacks well-defined representation and evaluation. 
Taking advantage of a super-sentential level of AMR coreference annotation from previous work,
we introduce a simple algorithm for deriving a unified graph representation,
avoiding the pitfalls of information loss from over-merging and lack of coherence from under-merging. 
Next, we describe improvements to the Smatch metric to make it tractable for comparing document-level graphs, 
and use it to re-evaluate the best published document-level AMR parser. We also present a pipeline approach combining the top performing AMR parser and coreference resolution systems, providing a strong baseline for future research.  
\end{abstract}

\section{Introduction}



Abstract Meaning Representation (AMR) is a formalism that represents the meaning of a text in the form of a directed graph, where nodes represent concepts and edges are labeled with relations \cite{banarescu-etal-2013-abstract}. Until recently, the annotated corpora for AMR only provided sentence-level annotations. As a result, AMR parsing research has been limited to sentence-level parsing. A unified document-level AMR graph representation can be quite useful for applications that require document-level understanding, such as question answering and summarization. 

The most recent release of AMR annotations (AMR 3.0) includes a multi-sentence AMR corpus \citep[MS-AMR;][]{ogorman-etal-2018-amr} that provides cross-sentential coreference information for AMR graphs. These annotations connect multiple sentence-level graphs via coreference chains, implicit relations and bridging relations. However, an adequate representation of this multi-sentence information in the form of an AMR graph has yet to be decided upon. The initial proposal \citep{ogorman-etal-2018-amr}, to merge all nodes in a coreference chain into one node, suffers from significant information loss. Without a consistent multi-sentence graph representation, AMR's standard Smatch metric \citep{smatch} cannot be used for evaluation and comparison. The Smatch scores differ greatly depending on how the coreference information is added to the AMR graphs. Additionally, the Smatch algorithm is exceedingly slow in its original form when run over multi-sentence graphs. These limitations in multi-sentence AMR graph representation and evaluation get in the way of research efforts in this area. Without a consistent representation and evaluation mechanism, it is impossible to draw meaningful comparisons between different approaches on the task. In this work, we present a simple and non-lossy multi-sentence AMR graph representation as well as a modification of the Smatch algorithm that allows efficient and consistent evaluation\footnote{The code for producing the proposed representation and efficient Smatch evaluation under Apache License 2.0 is available at: \url{https://github.com/IBM/docAMR}}.

\begin{figure*}[ht!]
    \centering


 \scalebox{0.8}{

\begin{tikzpicture}

\node (Bill) at (0.0,0) {\strut Bill};
\node [right = 4pt of Bill] (left) {\strut left};
\node [right = 4pt of left] (for) {\strut for};
\node [right = 4pt of for] (Paris) {\strut Paris};
\node [right = 4pt of Paris] (dot) {\strut .};
\node [right = 4pt of dot] (he) {\strut He};
\node [right = 4pt of he] (arrived) {\strut arrived};
\node [right = 4pt of arrived] (at) {\strut at};
\node [right = 4pt of at] (noon) {\strut noon};
\node [right = 4pt of noon] (dot2) {\strut .};

\node [draw,rounded corners, above = 110pt of left] (leave-11) {leave-11};
\node [draw,rounded corners, above = 70pt of Bill] (person) {person};
\node [draw,rounded corners, above = 40pt of Bill] (name_p) {name};
\node [draw,rounded corners, above = 10pt of Bill] (Bill) {Bill};

\node [draw,rounded corners, above = 70pt of Paris] (city) {city};
\node [draw,rounded corners, above = 40pt of Paris] (name_c) {name};
\node [draw,rounded corners, above = 10pt of Paris] (Paris) {Paris};

\node [draw,rounded corners, above = 70pt of he] (he) {he};
\node [draw,rounded corners, above = 110pt of arrived] (arrive-01) {arrive-01};
\node [draw,rounded corners, above = 70pt of noon] (noon) {noon};

\draw [-latex,thick] (leave-11) -- node[left,pos=0.5]  {\footnotesize :ARG0} (person);
\draw [-latex,thick] (person) -- node[left,pos=0.5]  {\footnotesize :name} (name_p);
\draw [-latex,thick] (name_p) -- node[left,pos=0.5]  {\footnotesize :op1} (Bill);

\draw [-latex,thick] (leave-11) -- node[left,pos=0.5]  {\footnotesize :ARG2} (city);
\draw [-latex,thick] (city) -- node[left,pos=0.5]  {\footnotesize :name} (name_c);
\draw [-latex,thick] (name_c) -- node[left,pos=0.5]  {\footnotesize :op1} (Paris);

\draw [-latex,thick] (arrive-01) -- node[left,pos=0.5]  {\footnotesize :ARG1} (he);
\draw [-latex,thick] (arrive-01) -- node[left,pos=0.5]  {\footnotesize :time} (noon);

\draw [densely dashed,thick,red] (person) to[out=-20,in=-150] node[right,pos=0.8,red]  {\footnotesize chain} (he);
\draw [-latex,densely dashed,thick,blue] (arrive-01) to[out=-140,in=90] node[above,pos=0.4,blue]  {\footnotesize :ARG4} (city);
\end{tikzpicture}

\begin{tikzpicture}

\node (Bill) at (0.0,0) {\strut Bill};
\node [right = 4pt of Bill] (left) {\strut left};
\node [right = 4pt of left] (for) {\strut for};
\node [right = 4pt of for] (Paris) {\strut Paris};
\node [right = 4pt of Paris] (dot) {\strut .};
\node [right = 4pt of dot] (he) {\strut He};
\node [right = 4pt of he] (arrived) {\strut arrived};
\node [right = 4pt of arrived] (at) {\strut at};
\node [right = 4pt of at] (noon) {\strut noon};
\node [right = 4pt of noon] (dot2) {\strut .};

\node [draw,rounded corners, above = 140pt of Paris] (doc) {document};

\node [draw,rounded corners, above = 110pt of left] (leave-11) {leave-11};
\node [draw,rounded corners, above = 70pt of Bill,red] (person) {person};
\node [draw,rounded corners, above = 40pt of Bill,red] (name_p) {name};
\node [draw,rounded corners, above = 10pt of Bill,red] (Bill) {Bill};

\node [draw,rounded corners, above = 70pt of Paris,blue] (city) {city};
\node [draw,rounded corners, above = 40pt of Paris,blue] (name_c) {name};
\node [draw,rounded corners, above = 10pt of Paris,blue] (Paris) {Paris};

\node [draw,rounded corners, above = 40pt of he,red] (he) {he};
\node [draw,rounded corners, above = 110pt of arrived] (arrive-01) {arrive-01};
\node [draw,rounded corners, above = 70pt of noon] (noon) {noon};

\draw [-latex,thick] (leave-11) -- node[left,pos=0.5]  {\footnotesize :ARG0} (person);
\draw [-latex,thick,red] (person) -- node[left,pos=0.5,red]  {\footnotesize :name} (name_p);
\draw [-latex,thick,red] (name_p) -- node[left,pos=0.5,red]  {\footnotesize :op1} (Bill);

\draw [-latex,thick] (leave-11) -- node[left,pos=0.5]  {\footnotesize :ARG2} (city);
\draw [-latex,thick,blue] (city) -- node[right,pos=0.5,blue]  {\footnotesize :name} (name_c);
\draw [-latex,thick,blue] (name_c) -- node[right,pos=0.5,blue]  {\footnotesize :op1} (Paris);

\draw [-latex,thick,blue] (arrive-01) -- node[right,pos=0.5,blue]  {\footnotesize :ARG4} (city);
\draw [-latex,thick] (arrive-01) -- node[above,pos=0.3]  {\footnotesize :ARG1} (person);
\draw [-latex,thick] (arrive-01) -- node[left,pos=0.5]  {\footnotesize :time} (noon);

\draw [-latex,thick,red] (person) -- node[right,pos=0.1,red]  {\footnotesize :instance} (he);

\draw [-latex,thick] (doc) -- node[left,pos=0.5]  {\footnotesize :snt1} (leave-11);
\draw [-latex,thick] (doc) -- node[left,pos=0.5]  {\footnotesize :snt2} (arrive-01);


\end{tikzpicture} 
 }

\caption{Two example sentences (on the left) annotated with cross sentential identity chain ({\color{red} red}) and implicit relation ({\color{blue} blue}) and the corresponding merged representation (on the right) as per \citep{ogorman-etal-2018-amr} }

    \label{fig1}

\end{figure*}
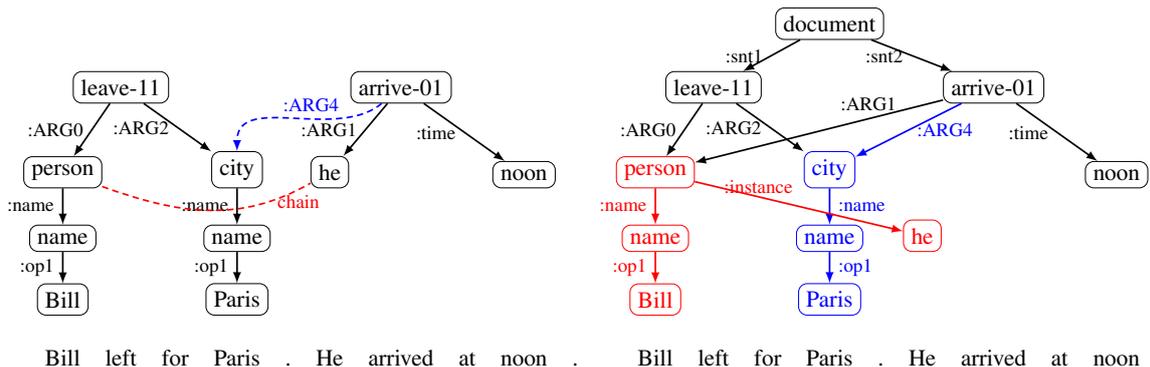

Our chief contributions are:
\begin{itemize}
    \item A standard for merging sentence-level AMR graphs into a single document-level graph based on MS-AMR coreference annotations (\repname; \cref{sec:docamr})
    \item A faster implementation of the Smatch metric suitable for evaluating document-level AMR parsers (\cref{sec:smatch})
    \item A metric, based on Smatch, for specific evaluation of coreference in document-level AMR graphs (\cref{sec:subscore})
    \item A baseline system for document-level AMR parsing and its evaluation (\cref{sec:parsing})
\end{itemize}

\section{Multi-Sentence AMR Corpus}

Multi-Sentence AMR corpus (henceforth MS-AMR) is part of AMR3.0 release and was introduced by \citet{ogorman-etal-2018-amr}. While a number of works have used document-level AMR graph creation for downstream tasks such as summarization \citep{liu-etal-2015-toward,lee2021amrsummarization}, it is the only dataset currently with manual annotations of AMR document graphs. It contains 284 documents in its train split and 9 documents on test split, annotated over 8027 gold AMRs in total. Out of the 284 train documents, 43 are annotated twice; in this work we refer to these double annotation documents as our development set. 

Coreference chains were annotated as clusters of coreferent nodes in gold AMR graphs, rather than as coreferent spans of text: this is what allows evaluation with Smatch. Such an annotation also precludes evaluation of system predictions using more traditional coreference evaluations, as there is no notion of mention spans to use when comparing mentions. 

The MS-AMR corpus also annotated implicit relations, as shown in the blue \amr{ARG4} edge in \cref{fig1}, whenever a numbered argument of a predicate was semantically identifiable in another sentence.  
Similarly, the MS-AMR corpus annotates bridging mentions, such as part-whole or set-member relations. Both implicit relations and bridging relations can be seamlessly added as cross-sentential edges.

Coreference chains form the majority of cross-sentential relations annotated in MS-AMR. Unlike implicit and bridging relations, there is no straight forward way of incorporating coreference chains into a multi-sentence AMR graph. Next section discusses the challenges involved and proposes a solution. 

\section{Document-level AMR Representation}

\begin{figure*}[!ht]
    \centering

 \scalebox{0.8}{
 
\begin{tikzpicture}


\begin{scope}
\node (John) at (0,5) {\strut John's};
\node [right = 3pt of John] (hatred) {\strut hatred of};
\node [right = 3pt of hatred] (cats1) {\strut cats};
\node [right = 3pt of cats1] (dots) {\strut ...};

\node [right = 3pt of dots] (he) {\strut he};
\node [right = 3pt of he] (like) {\strut doesn't like};
\node [right = 3pt of like] (them) {\strut them};
\node [right = 3pt of them] (dots) {\strut ...};


\node [draw,rounded corners, teal, above = 70pt of John] (person) {\strut person};
\node [draw,rounded corners, teal, above = 40pt of John] (name) {\strut name};
\node [draw,rounded corners, teal, above = 10pt of John] (John) {\strut John};
\node [draw,rounded corners, teal, above = 70pt of cats1] (cats1n) {\strut cats};
\node [draw,rounded corners, teal, above = 120pt of hatred] (hate-01) {\strut hate-01};


\node [draw,rounded corners, violet, above = 70pt of he] (he) {\strut he};
\node [draw,rounded corners, violet, above = 70pt of them] (they) {\strut they};
\node [draw,rounded corners, violet, above = 70pt of like] (-) {\strut -};
\node [draw,rounded corners, violet, above = 120pt of like] (like-01) {\strut like-01};

\draw [-latex,thick, teal] (person) -- node[left,pos=0.5]  {\footnotesize} (name);
\draw [-latex,thick, teal] (name) -- node[left,pos=0.5]  {\footnotesize } (John);
\draw [-latex,thick, teal] (hate-01) -- node[left,pos=0.5]  {\footnotesize :ARG0} (person);
\draw [-latex,thick, teal] (hate-01) -- node[right,pos=0.5]  {\footnotesize :ARG1} (cats1n);

\draw [-latex,thick,violet] (like-01) -- node[left,pos=0.5]  {\footnotesize :ARG0} (he);
\draw [-latex,thick,violet] (like-01) -- node[right,pos=0.5]  {\footnotesize :ARG1} (they);
\draw [-latex,thick, violet] (like-01) -- node[pos=0.5]  {\footnotesize :polarity} (-);
\draw [densely dashed,thick] (like-01) to[out=-170,in=-10] node[above,pos=0.5]  {\footnotesize chain} (hate-01);
\draw [densely dashed,thick] (person) to[out=-20,in=-160] node[above,pos=0.4]  {\footnotesize chain} (he);
\draw [densely dashed,thick] (cats1n) to[out=-20,in=-160] node[above,pos=0.5]  {\footnotesize chain} (they);

\end{scope}


\begin{scope}
\node (John) at (10,5) {\strut John's};
\node [right = 3pt of John] (hatred) {\strut hatred of};
\node [right = 3pt of hatred] (cats1) {\strut cats};
\node [right = 3pt of cats1] (dots) {\strut ...};

\node [right = 3pt of dots] (he) {\strut he};
\node [right = 3pt of he] (like) {\strut doesn't like};
\node [right = 3pt of like] (them) {\strut them};
\node [right = 3pt of them] (dots) {\strut ...};


\node [draw,rounded corners, teal, above = 70pt of John] (person) {\strut person};
\node [draw,rounded corners, teal, above = 40pt of John] (name) {\strut name};
\node [draw,rounded corners, teal, above = 10pt of John] (John) {\strut John};
\node [draw,rounded corners, teal, above = 70pt of cats1] (cats1n) {\strut cats};
\node [draw,rounded corners, teal, above = 120pt of hatred] (hate-01) {\strut hate-01};


\node [draw,rounded corners, violet, above = 70pt of he] (he) {\strut he};
\node [draw,rounded corners, violet, above = 70pt of them] (they) {\strut they};
\node [draw,rounded corners, violet, above = 70pt of like] (-) {\strut -};
\node [draw,rounded corners, violet, above = 120pt of like] (like-01) {\strut like-01};

\draw [-latex,thick] (person) -- node[left,pos=0.5]  {\footnotesize} (name);
\draw [-latex,thick] (name) -- node[left,pos=0.5]  {\footnotesize } (John);
\draw [-latex,thick] (hate-01) -- node[left,pos=0.5]  {\footnotesize :ARG0} (person);
\draw [-latex,thick] (hate-01) -- node[right,pos=0.5]  {\footnotesize :ARG1} (cats1n);

\draw [-latex,thick] (hate-01) -- node[right, pos=0.5]  {\footnotesize :polarity} (-);
\draw [-latex,thick] (hate-01) -- node[above,pos=0.5]  {\footnotesize :instance} (like-01);
\draw [-latex,thick] (person) to[out=-30,in=-150] node[above,pos=0.4]  {\footnotesize :instance} (he);
\draw [-latex,thick] (cats1n) to[out=-30,in=-150] node[above,pos=0.5]  {\footnotesize :instance} (they);
\end{scope}


\begin{scope}
\node (favor) at (0,0) {\strut favor};
\node [right = 3pt of favor] (dots1) {\strut ...};

\node [right = 3pt of dots1] (give) {\strut give};
\node [right = 3pt of give] (him) {\strut him};
\node [right = 3pt of him] (a) {\strut a};
\node [right = 3pt of a] (lift) {\strut lift};
\node [right = 3pt of lift] (dots2) {\strut ...};

\node [right = 3pt of dots2] (help-out) {\strut help out};
\node [right = 3pt of help-out] (a) {\strut a};
\node [right = 3pt of a] (fellow) {\strut fellow};

\node [draw,rounded corners, red, above = 10pt of favor] (favor) {\strut favor};

\node [draw,rounded corners,teal, above = 70pt of give] (give-01) {\strut give-01};
\node [draw,rounded corners,teal, above = 10pt of give] (you) {\strut you};
\node [draw,rounded corners,teal, above = 10pt of lift] (lift) {\strut lift};
\node [draw,rounded corners,teal, above = 10pt of him] (he) {\strut he};

\node [draw,rounded corners,violet, above = 70pt of help-out] (help-01) {\strut help-01};
\node [draw,rounded corners,violet, above = 10pt of help-out] (out) {\strut out};
\node [draw,rounded corners,violet, above = 10pt of fellow] (fellow) {\strut fellow};

\draw [-latex,thick,teal] (give-01) -- node[pos=0.6]  {\footnotesize :ARG0} (you);
\draw [-latex,thick,teal] (give-01) -- node[right,pos=0.7]  {\footnotesize :ARG2} (he);
\draw [-latex,thick,teal] (give-01) -- node[right,pos=0.7]  {\footnotesize :ARG1} (lift);

\draw [-latex,thick,violet] (help-01) -- node[right,pos=0.6]  {\footnotesize :ARG2} (fellow);
\draw [-latex,thick,violet] (help-01) -- node[left,pos=0.6]  {\footnotesize :manner} (out);

\draw [densely dashed,thick] (favor) to[out=90,in=-140] node[above,pos=0.5]  {\footnotesize chain} (give-01);
\draw [densely dashed,thick] (give-01) to[out=-10,in=-170] node[above,pos=0.5]  {\footnotesize chain} (help-01);
\draw [densely dashed,thick] (he) to[out=-20,in=-160] node[above,pos=0.5]  {\footnotesize chain} (fellow);
\end{scope}


\begin{scope}

\node (favor) at (10,0) {\strut favor};
\node [right = 3pt of favor] (dots1) {\strut ...};

\node [right = 3pt of dots1] (give) {\strut give};
\node [right = 3pt of give] (him) {\strut him};
\node [right = 3pt of him] (a) {\strut a};
\node [right = 3pt of a] (lift) {\strut lift};
\node [right = 3pt of lift] (dots2) {\strut ...};

\node [right = 3pt of dots2] (help-out) {\strut help out};
\node [right = 3pt of help-out] (a) {\strut a};
\node [right = 3pt of a] (fellow) {\strut fellow};

\node [draw,rounded corners, red, above = 10pt of favor] (favor) {\strut favor};

\node [draw,rounded corners,teal, above = 70pt of give] (give-01) {\strut give-01};
\node [draw,rounded corners,teal, above = 10pt of give] (you) {\strut you};
\node [draw,rounded corners,teal, above = 10pt of lift] (lift) {\strut lift};
\node [draw,rounded corners,teal, above = 10pt of him] (he) {\strut he};

\node [draw,rounded corners,violet, above = 70pt of help-out] (help-01) {\strut help-01};
\node [draw,rounded corners,violet, above = 10pt of help-out] (out) {\strut out};
\node [draw,rounded corners,violet, above = 10pt of fellow] (fellow) {\strut fellow};

\draw [-latex,thick] (favor) to[out=35,in=145] node[above,pos=0.9]  {\footnotesize :ARG0} (you);
\draw [-latex,thick] (favor) to[out=35,in=145] node[above,pos=0.9]  {\footnotesize :ARG2} (he);
\draw [-latex,thick] (favor) to[out=35,in=145] node[above,pos=0.9]  {\footnotesize :ARG1} (lift);

\draw [-latex,thick] (favor) to[out=35,in=145] node[above,pos=0.9]  {\footnotesize :ARG2} (fellow);
\draw [-latex,thick] (favor) to[out=35,in=145] node[above,pos=0.8]  {\footnotesize :manner} (out);

\draw [-latex,thick] (favor) to[out=35,in=-150] node[left,pos=0.5]  {\footnotesize :instance} (give-01);
\draw [-latex,thick] (favor) to[out=35,in=155] node[right,pos=0.6]  {\footnotesize :instance} (help-01);
\draw [-latex,thick] (he) to[out=15,in=90] node[above,pos=0.7]  {\footnotesize :instance} (fellow);

\end{scope}

\end{tikzpicture}

}

\caption{Two examples of merge operations that result in loss or distortion of information. Nodes of the same color belong to the same sentence-level AMR. Dashed lines indicate the identity chain links. In the top example, the merged node representation (right) reverses the meaning of the first sentence because of the negative \amr{:polarity} node migrating from the merged node \amr{like-01}. In the bottom example, the non-predicate node \amr{favor} has received a bunch of \amr{ARG} nodes from the fellow coreference nodes. Moreover, the merged-in predicate nodes \amr{give-01} and \amr{help-01} have lost association with their arguments. For example, the sentence "you help the fellow lift [something]" will be consistent with this graph. }

    \label{fig3}

\end{figure*}
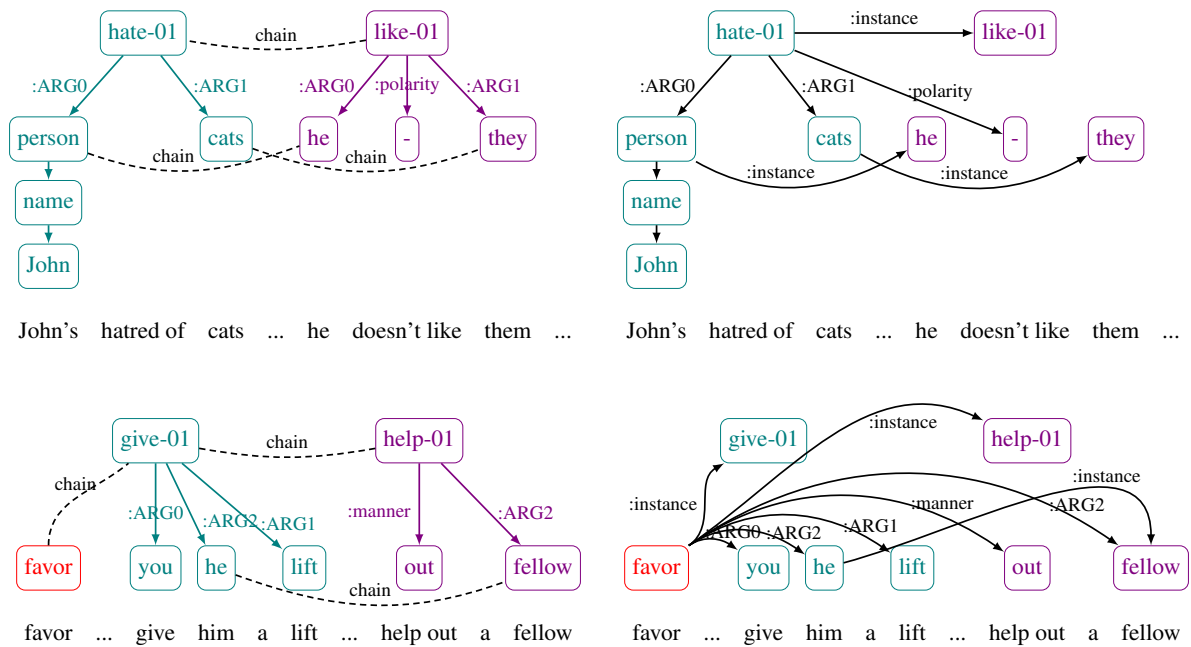

As discussed in the past section, majority of MS-AMR annotations take the form of coreference chains over sentence-level gold AMR graphs.  In this work, we propose a representation \repname that incorporates these chains into multi-sentence AMR graphs. The motivation for having such a representation is two-fold.
First,  representing a multi-sentence document as a single AMR graph will allow downstream applications to treat the whole document as single entity, much like the AMR graphs for sentences. 
Second, it is crucial to have a consistent representation for the purpose of evaluating MS-AMR predictions. 
Smatch -- the standard evaluation metric for AMR -- operates over graphs. We can use this metric for MS-AMR evaluation only if the system outputs and the gold graphs adhere to the same document-level graph representation. 

\textbf{Intrasentential coreferences} in \emph{sentence-level} AMR are annotated with what one might call \emph{manual full merge} -- i.e., given multiple mentions in the same sentence (such as a named entity and a pronoun), a human annotator would pick a single informative mention, and merge all other information into the root of the subgraph representing that mention. Language being efficient, it is very rare for multiple non-reduced mentions to occur in the same sentence, therefore it was a trivial task for human annotators. MS-AMR human annotations do not perform manual merge -- they only indicate the nodes in multiple existing sentence-level graphs that participate in a coreference cluster. 

\subsection{Why Not Merge All Coreferent Nodes?}



The pioneering work on MS-AMR \citep{ogorman-etal-2018-amr} that introduced the annotations, also proposed a way to represent MS-AMR in the form of a graph.  This approach essentially merges all coreferent nodes from a chain into one node that inherits all the parents and children of the original nodes.  Furthermore, if the lexical form of the merged nodes are different, each distinct form is added as an additional instance of the merged node connected via an \amr{:instance} edge. \Cref{fig1} shows a simple example of this merge operation. 

This representation has a desirable property in that it treats cross sentential coreference information in a manner similar to sentence-level reentrancies. Another advantage of node merging is that a single node represents each entity/event, thus a document representation can be viewed as a mini Knowledge Base for the document.  Moreover, this highly connected representation is consequential for Smatch: if a gold document has a cluster of ten mentions and a system incorrectly splits that cluster into six and four mentions, then the optimal one-to-one Smatch alignment will consider all links to the smaller cluster wrong. 

However, merging coreference nodes at the document level is not trivial. Automatic indiscriminate merging of all coreference clusters has serious limitations. Without manual curation of potentially conflicting referential expressions or situations, it can end up merging conflicting information, particularly with event coreference: e.g., ``his hatred of cats'' and ``John doesn't like cats'' would be merged into the AMR equivalent of ``John+he does not (like + hate) cats''.  It does not preserve the semantics of the original AMR graphs. \Cref{fig3} shows examples of merge operations that either lose or distort the meaning of the text. 

    Moreover, the proposed method of \citep{ogorman-etal-2018-amr} assumes that a node can have multiple \amr{:instance} edges -- however existing AMR tools (including the Smatch implementation) do not allow multiple \amr{:instance} edges on a node. In the hand full of previous works on MS-AMR, edge labels other than \amr{:instance} are used to connect additional forms of a node. One node becomes privileged over others under the proposed approach -- as a result, a system that selects the privileged node differently will be unnecessarily penalized.


We wish to avoid the above pitfalls while still retaining the advantages of node merging. The next section discusses the principles considered while formulating the \repname representation. 

\subsection{Considerations}
\label{sec:considerations}

Our primary consideration while designing \repname has been to preserve the meaning of the underlying text as encoded by the combination of sentence-level graphs and MS-AMR annotations. Whenever a secondary consideration led towards a lossy representation, we decided in favor of meaning preservation. In addition, we considered the following principles:

\begin{enumerate}
    \item \textit{Be consistent with AMR's treatment of within-sentence coreference.} In sentence-level AMRs, pronominal concepts are only used in the absence of a contentful antecedent. Pronouns don’t contribute enough meaning beyond referential information. Therefore, only those pronouns are kept whose antecedent is unspecified. Furthermore, in sentence-level AMRs all coreferent nodes are merged so they share the same variable. 
    We stick to these principles except in the cases where they can potentially cause information loss.
    \item \textit{Do not place more weight on any particular representative member of a coreference chain.} It would be artificial to represent the cluster as pairwise mention-mention links, e.g.~by privileging one mention as the primary representative of the cluster. Instead, coref chains should be represented in a way that the non-pronominal ways of referring to the entity/event in different sentences are on equal footing.
    \item \textit{Systems should not get too much credit for superficial decisions} (like merely creating a concept for a pronoun, or detecting the name string and entity type on every mention of the name). This can be achieved by merging the nodes whenever it can be done safely without loss of information. The effect is that the scoring places more weight on detection of coreference.
    \item \textit{The representation should support comparison via Smatch} (without unnecessary computational overhead) for document-level parsers given the same document as input.\footnote{We do not consider here how to measure similarity between \emph{comparable} documents.}
    \item \textit{The edges between the nodes in a coref cluster and their sentence-level neighbours should be kept intact}, so that when evaluated with Smatch, at least one mention in a chain of correctly predicted entity mentions will get full credit even if the coreference is missed. 
    \item Finally, \textit{the representation should not rely on complicated heuristics or new manual annotations}. For example, we considered and ultimately rejected the possibility of merging coreferent non-name nodes, which would have required heuristics for choosing one representative concept from among multiple distinct concepts from different mentions (and also would have violated principle \#2).

\end{enumerate}
\subsection{Proposed Approach (\repname)}\label{sec:docamr}


We propose \repname, an extension of AMR to the document level. In this representation, the entity and event coreference annotations from  \citep{ogorman-etal-2018-amr} are properly assimilated into a single graph representing the meaning of the document. It completely preserves the semantic information while trying to stay close to  the sentence-level AMR conventions.

\begin{figure}[!t]
    \centering

 \scalebox{0.75}{
 
\begin{tikzpicture}
\begin{scope}
\node (favor) at (0,6) {\strut favor};
\node [right = 3pt of favor] (dots1) {\strut ...};

\node [right = 3pt of dots1] (give) {\strut give};
\node [right = 3pt of give] (him) {\strut him};
\node [right = 3pt of him] (lift) {\strut a lift};
\node [right = 3pt of lift] (dots2) {\strut ...};

\node [right = 3pt of dots2] (help-out) {\strut help out};
\node [right = 3pt of help-out] (a) {\strut a};
\node [right = 3pt of a] (fellow) {\strut fellow};
\node [right = 3pt of fellow] (dots3) {\strut ...};

\node [draw,rounded corners, above = 90pt of favor] (coref-entity1) {\strut \textbf{coref-entity}};
\node [draw,rounded corners, red, above = 10pt of favor] (favor) {\strut favor};

\node [draw,rounded corners,teal, above = 70pt of give] (give-01) {\strut give-01};
\node [draw,rounded corners,teal, above = 10pt of give] (you) {\strut you};
\node [draw,rounded corners,teal, above = 10pt of lift] (lift) {\strut lift};
\node [draw,rounded corners, dotted, text=gray, above = 10pt of him] (he) {\strut he};

\node [draw,rounded corners,violet, above = 70pt of help-out] (help-01) {\strut help-01};
\node [draw,rounded corners,violet, above = 10pt of help-out] (out) {\strut out};
\node [draw,rounded corners,violet, above = 10pt of fellow] (fellow) {\strut fellow};


\draw [-latex,thick,dotted,gray] (give-01) -- node[pos=0.7]  {\footnotesize :ARG2} (he);
\draw [-latex,thick,teal] (give-01) -- node[pos=0.5]  {\footnotesize :ARG0} (you);
\draw [-latex,thick,teal] (give-01) -- node[right,pos=0.3]  {\footnotesize :ARG2} (fellow);
\draw [-latex,thick,teal] (give-01) -- node[right,pos=0.5]  {\footnotesize :ARG1} (lift);

\draw [-latex,thick,violet] (help-01) -- node[right,pos=0.5]  {\footnotesize :ARG2} (fellow);
\draw [-latex,thick,violet] (help-01) -- node[pos=0.6]  {\footnotesize :manner} (out);

\draw [latex-,,thick] (favor) -- node[above,pos=0.8]  {\footnotesize :coref} (coref-entity1);
\draw [latex-,,thick] (give-01) to[out=90,in=45] node[above,pos=0.5]  {\footnotesize :coref} (coref-entity1);
\draw [latex-,thick] (help-01) to[out=180,in=55] node[above,pos=0.5]  {\footnotesize :coref} (coref-entity1);


\end{scope}

\end{tikzpicture}

}

\caption{Our proposed \repname~representation applied to the bottom example from Figure \ref{fig3}. For the identity chain between \amr{he} and \amr{fellow}, the pronoun is dropped indicated in dotted gray and the links are transferred to the non-pronominal node. For the identity chain between \amr{favor}, \amr{give-01} and \amr{help-01}, merging is deemed potentially lossy -- instead a \amr{coref-entity} node is introduced and all nodes in the chain are linked to it via \amr{:coref} edge. This not only preserves the  meaning without loss but also avoids preferential treatment of any content node in the chain.
}
    \label{fig4}

\end{figure}
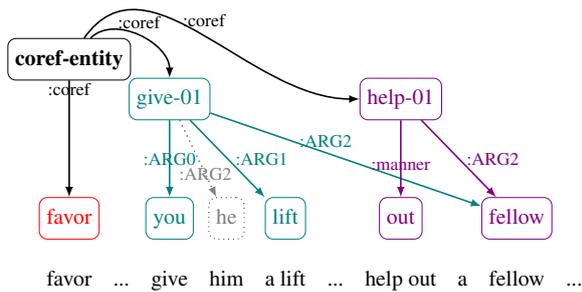

The basic idea is that a new node with concept \amr{coref-entity} is added for each identity chain of nodes -- all nodes participating in the chain are then connected to this node via a \amr{:coref} relation. \Cref{fig4} shows an example of this. In addition to the base method of using \amr{coref-entity} nodes, we introduce two exceptions. First, all the \textbf{named entities} within an identity chain are merged into one named entity. Second, all \textbf{pronominal nodes} participating in a chain are dropped. Both these exceptions are consistent with the treatment of coreferent named entities and pronouns in sentence-level AMR annotations. Further details of these two exceptions are outlined below:

\paragraph{Treatment of Named Entities:} All the named entities (that is, AMR nodes with a \amr{:name} relation) participating in an identity chain cluster are merged into one named entity. The structure of named entities is quite consistent and usually all the named entities participating in a chain match exactly. Occasionally, a cluster may contain named entities that differ in their type, wiki link or form of the name. In such cases, all unique forms of the name and all unique wikis are kept under a common variable. If types are different, the most specific type becomes the root and remaining types are connected to it with the new relation  \amr{:additional-type}.\footnote{We use the AMR types ontology to decide which type is the most specific. If a cluster contains types that are not in the AMR ontology, then the ones from the ontology are considered for root position; and if none belong to the ontology, the most frequent type from the within the cluster becomes the root. Frequency ties are broken by picking the first mention.} If nodes to be merged have modifier roles other than \amr{:name} and \amr{:wiki}, they are gathered under the merged entity node.

\paragraph{Treatment of Pronominal Nodes:} 
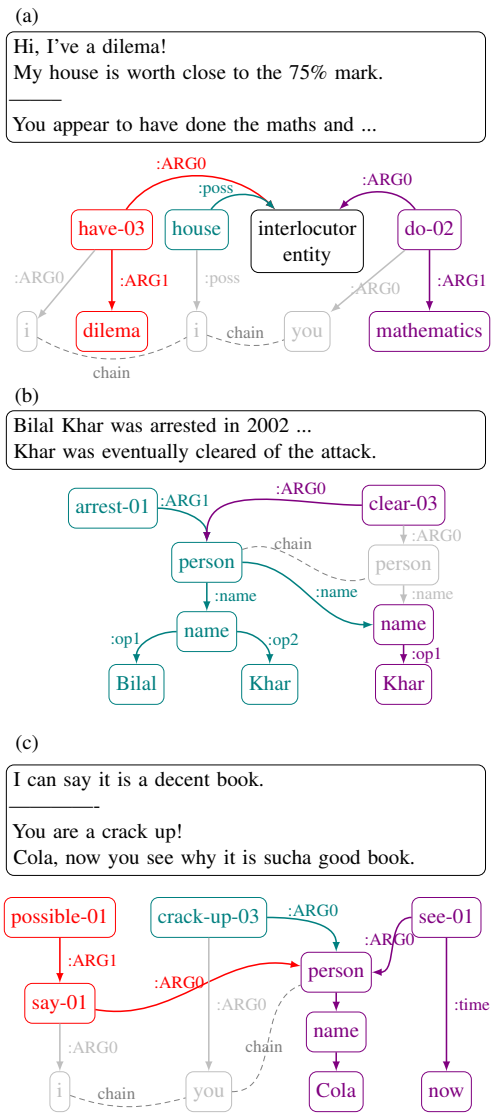
\begin{figure}[ht!]
    \centering

 \scalebox{0.72}{
 
\begin{tikzpicture}
\begin{scope}

\node (lab_c) at (0,6.4) {\strut (c)};
\node (lab_c) at (0,12.8) {\strut (b)};
\node (lab_c) at (0,19.8) {\strut (a)};


\node [draw,rounded corners, lightgray] (i) at (0.6,0) {\strut i};
\node [draw,rounded corners, red, above = 25pt of i] (say-01) {\strut say-01};
\node [draw,rounded corners, red, above = 70pt of i] (possible-01) {\strut possible-01};

\node [draw,rounded corners, lightgray, right = 60pt of i] (you) {\strut you};
\node [draw,rounded corners,teal, above = 70pt of you] (crack-up-03) {\strut crack-up-03};

\node [draw,rounded corners, violet, right = 40pt of you] (Cola) {\strut Cola};
\node [draw,rounded corners, violet, above = 10pt of Cola] (name) {\strut name};
\node [draw,rounded corners, violet, above = 10pt of name] (person) {\strut person};

\node [draw,rounded corners,violet, right = 30pt of Cola] (now) {\strut now};
\node [draw,rounded corners,violet, above = 70pt of now] (see-01) {\strut see-01};

\draw [-latex,thick,red] (possible-01) -- node[right,pos=0.5]  {\footnotesize :ARG1} (say-01);
\draw [-latex,thick,lightgray] (say-01) -- node[right,pos=0.5]  {\footnotesize :ARG0} (i);

\draw [-latex,thick,red] (say-01) to[out=-10,in=170] node[above,pos=0.4]  {\footnotesize :ARG0} (person);
\draw [-latex,thick,lightgray] (crack-up-03) -- node[right,pos=0.5]  {\footnotesize :ARG0} (you);
\draw [-latex,thick,teal] (crack-up-03) to[out=0,in=90] node[above,pos=0.5]  {\footnotesize :ARG0} (person);

\draw [-latex,thick, violet] (person) -- node[left,pos=0.5]  {\footnotesize} (name);
\draw [-latex,thick, violet] (name) -- node[left,pos=0.5]  {\footnotesize } (Cola);

\draw [-latex,thick,violet] (see-01) to[out=180,in=0] node[above,pos=0.6]  {\footnotesize :ARG0} (person);
\draw [-latex,thick,violet] (see-01) -- node[right,pos=0.5]  {\footnotesize :time} (now);

\draw [densely dashed,gray] (i) to[out=-20,in=-165] node[above,pos=0.4]  {\footnotesize chain} (you);
\draw [densely dashed,gray] (you) to[out=0,in=-160] node[above,pos=0.4]  {\footnotesize chain} (person);


\node [draw,rounded corners, align=left,text width=8.5cm] (text) at (4,5) {I can say it is a decent book.\\-------------\\You are a crack up!\\Cola, now you see why it is sucha good book.};

\end{scope}


\begin{scope}

\node [draw,rounded corners, teal] (Bilal) at (2,7.5) {\strut Bilal};
\node [draw,rounded corners, teal, right = 40pt of Bilal] (Khar) {\strut Khar};
\node [draw,rounded corners, teal, above right = 9pt of Bilal] (name) {\strut name};
\node [draw,rounded corners, teal, above = 15pt of name] (person) {\strut person};
\node [draw,rounded corners, teal, above left = 10pt of person, above left = 10pt of person] (arrest-01) {\strut arrest-01};


\node [draw,rounded corners, violet, right = 40pt of Khar] (Khar2) {\strut Khar};
\node [draw,rounded corners, violet, above = 10pt of Khar2] (name2) {\strut name};
\node [draw,rounded corners, lightgray, above = 10pt of name2] (person2) {\strut person};
\node [draw,rounded corners, violet, above = 10pt of person2] (clear-03) {\strut clear-03};

\draw [-latex,thick, teal] (arrest-01) to[out=0,in=90] node[above,pos=0.4]  {\footnotesize :ARG1} (person);
\draw [-latex,thick, teal] (person) -- node[right,pos=0.5]  {\footnotesize :name} (name);
\draw [-latex,thick, teal] (name) to[out=180,in=90] node[left,pos=0.5]  {\footnotesize :op1} (Bilal);
\draw [-latex,thick, teal] (name) to[out=0,in=90] node[right,pos=0.5]  {\footnotesize :op2} (Khar);
\draw [-latex,thick, teal] (person) to[out=0,in=180] node[right,pos=0.5]  {\footnotesize :name} (name2);


\draw [-latex,thick, lightgray] (clear-03) -- node[right,pos=0.5]  {\footnotesize :ARG0} (person2);
\draw [-latex,thick, lightgray] (person2) -- node[right,pos=0.5]  {\footnotesize :name} (name2);
\draw [-latex,thick, violet] (name2) -- node[right,pos=0.5]  {\footnotesize :op1} (Khar2);
\draw [-latex,thick, violet] (clear-03) to[out=180,in=90] node[above,pos=0.3]  {\footnotesize :ARG0} (person);

\draw [densely dashed,gray] (person) to[out=20,in=-160] node[above,pos=0.4]  {\footnotesize chain} (person2);


\node [draw,rounded corners, align=left,text width=8.5cm] (text) at (4,12) {Bilal Khar was arrested in 2002 ... \\Khar was eventually cleared of the attack.};

\end{scope}


\begin{scope}

\node [draw,rounded corners, lightgray] (i) at (0,14) {\strut i};
\node [draw,rounded corners, red, right = 20pt of i] (dilema) {\strut dilema};
\node [draw,rounded corners, red, above= 32pt of dilema] (have-03) {\strut have-03};

\node [draw,rounded corners, lightgray, right = 20pt of dilema] (i2) {\strut i};
\node [draw,rounded corners, teal, above = 32pt of i2] (house) {\strut house};


\node [draw,rounded corners, lightgray, right = 40pt of i2] (you) {\strut you};
\node [draw,rounded corners, violet, right = 20pt of you] (mathematics) {\strut mathematics};
\node [draw,rounded corners, violet, above = 32pt of mathematics] (do-02) {\strut do-02};
\node [draw,rounded corners,align=center, above= 20pt of you] (ile) {\strut interlocutor\\entity};

\draw [-latex,thick, lightgray] (house) -- node[right,pos=0.5]  {\footnotesize :poss} (i2);
\draw [-latex,thick, red] (have-03) to[out=55,in=135] node[above,pos=0.4]  {\footnotesize :ARG0} (ile);
\draw [-latex,thick, teal] (house) to[out=55,in=135] node[above,pos=0.1]  {\footnotesize :poss} (ile);
\draw [-latex,thick, violet] (do-02) to[out=125,in=45] node[above,pos=0.4]  {\footnotesize :ARG0} (ile);
\draw [-latex,thick, lightgray] (do-02) -- node[below,pos=0.4]  {\footnotesize :ARG0} (you);
\draw [-latex,thick, violet] (do-02) -- node[right,pos=0.5]  {\footnotesize :ARG1} (mathematics);

\draw [-latex,thick, lightgray] (have-03) -- node[left,pos=0.4]  {\footnotesize :ARG0} (i);
\draw [-latex,thick, red] (have-03) -- node[right,pos=0.5]  {\footnotesize :ARG1} (dilema);

\draw [densely dashed,gray] (i) to[out=-30,in=-150] node[below,pos=0.5]  {\footnotesize chain} (i2);
\draw [densely dashed,gray] (i2) to[out=-20,in=-160] node[above,pos=0.5]  {\footnotesize chain} (you);


\node [draw,rounded corners, align=left,text width=8.5cm] (text) at (4,18.5) {Hi, I've a dilema!\\My house is worth close to the 75\% mark.\\--------\\You appear to have done the maths and ...};

\end{scope}

\end{tikzpicture}

}

\caption{Various scenarios of merge operations in \repname: (a) Merging first and second person pronouns in a dialogue into \amr{interlocutor-entity} (b) Merging named entities with multiple forms, keeping all distinct forms under separate \amr{name} nodes (c) Dropping pronominal nodes and replacing them with the named entity in the chain Nodes of the same color belong to the same sentence-level AMR. Dotted gray indicates dropped nodes and edges.}


    \label{fig5}

\end{figure}
Sentential AMR incorporates pronouns as concepts only as a fallback if there is no contentful antecedent within the same sentence. 
We extend this philosophy to pronominal nodes whose antecedent can be found in another sentence.
That is, all pronominal nodes participating in an identity chain cluster with an antecedent are replaced by the \amr{chain-entity} node, removing the pronominal concepts from the graph. 
Pronominal concepts are retained only if there is no content antecedent to be found.
For an identity chain with exclusively pronominal nodes, no \amr{chain-entity} node is needed; they are simply merged into one node, which is labeled with the most specific pronoun concept in the chain (e.g., ``he'' is considered more specific than ``someone''). As a special case, if a heterogeneous pronoun chain (with multiple pronouns) refers to a participant in a dialogue -- indicated by the concept \amr{i} or \amr{you} -- a new \amr{interlocutor-entity} node is introduced and all pronouns are merged into that to account for different perspectives taken in different utterances. 

A notable consequence of discarding pronoun concepts is that, correct antecedent resolution is required in order for any roles in which the pronominal mention participates, to be counted as correct. Systems could struggle with long chains of pronouns, especially if the first in the chain is resolved to the wrong antecedent and this causes cascading errors. That said, we believe representing the wrong substantive entity as participating in a relation is a serious error, no less so when a system correctly clusters a group of pronouns but fails to resolve their antecedent correctly.

\paragraph{Discard Single Node Clusters:} After creating a document-level graph per the rules outlined above, we perform a final step of removing any \amr{coref-entity} nodes with one \amr{:coref} edge, and replace references to that node with its only member node. This can happen if there is a cluster with one non-pronominal concept and one or more pronouns that got deleted. It can also happen if a cluster had multiple names that were merged. This simplification will discard superfluous \amr{coref-entity} nodes. It would make alignment faster for Smatch that uses surface form matching during initialization.

\begin{table}[t]
\begin{tabular}{lccc}
\toprule 
    & Train    & Dev1 & Dev2 \\
\midrule 
Nodes in chains                    & 12513 & 1241 & 1141 \\
\midrule
Pronouns in chains                & 4456 & \hphantom{0}433 & \hphantom{0}423\\
$\hookrightarrow$ Pronoun             & 1029 & \hphantom{0}141 & \hphantom{0}121 \\
$\hookrightarrow$  Interlocutor-entity         & \hphantom{0}773  & \hphantom{00}66  & \hphantom{00}90 \\
$\hookrightarrow$  Other node & 1494 & \hphantom{0}140 & \hphantom{0}118  \\
$\hookrightarrow$  Coref-entity          & 1160 & \hphantom{00}86  & \hphantom{00}94   \\
\midrule
NEs in chains          & 2917 & \hphantom{0}224 & \hphantom{0}205 \\
NEs merged             & \hphantom{0}903  & \hphantom{00}85  & \hphantom{00}84 \\

\bottomrule
\end{tabular}

\caption{Statistics on \repname merge operations. Top row shows the total number of un-merged nodes in all identity-chains. Middle section shows the total number of pronominal nodes in identity-chain and the numbers of different target node types they are merged into. Most pronouns are dropped only about one fourth -- in pronoun-only chains -- are merged into one of the pronouns in the chain. Bottom sections shows number of Named Entities in chains before and after merging.}
    \label{tab:stats}

\end{table}





\subsection{The Roads not Taken}


In an attempt to adhere to the sentence-level AMR conventions (principle 1 in  \cref{sec:considerations}) to a greater extent, we considered merging all non-predicate nodes participating in a cluster -- keeping the most specific one. This required heuristics and even manual input on deciding the most specific instances, we therefore did not take that route. 

A less complicated version of this would have been merging only those nodes that have the same concept. However, even if a concept recurs, the different mentions may have different modifiers/arguments that may collide and make merging problematic. 
Merging only concepts with the same modifiers could place outsize importance on the attachment of modifiers by the parser. We opted for the simplest approach of not performing any merging of non-name, non-pronominal nodes.
  
We also considered an alternative representation where, similar to \repname, all nodes participating in a coreference cluster are connected to a \amr{coref-entity} -- but unlike \repname, the parents of all participating nodes now point to the \amr{coref-entity}. In other words, all mention nodes are clustered (via \amr{:coref} edge without any merge) under \amr{coref-entity} along with their descendent sub-graphs, but the \amr{coref-entity} nodes sit between the mention nodes and their sentence-level parents. This flipped version of \repname makes dense connections at document level making Smatch more sensitive to coreference errors. However, this often results in multiple similar or identical sub-graphs collected under \amr{coref-entity} -- with no connection to the corresponding sentences. Merging these sub-graphs based on the similarity of their structure will make the final representation highly dependent on small modifier level differences. Another significant side effect of this approach is that due to lack of connection with original sentence level graphs, the Smatch algorithm (as given in \cref{sec:smatch}) cannot benefit from sentence alignments and becomes prohibitively inefficient.

\section{\repname Evaluation}


\repname represents MS-AMR annotations of multiple sentences in the form of one AMR graph. Ideally, the quality of this graph should be assessed as single unified entity. 
Traditional measures of coreference, such as MUC, CEAF etc., try to align the gold coreference chains with the predicted ones based on the shared mentions. In the case of text based coreference resolution, identifying shared mentions is trivial since the mentions are anchored in the input text. AMR, by contrast, is not anchored in the text -- the nodes of AMR graphs are not aligned to words. As a result, traditional coreference measures can be applied to AMR only if the graphs are either identical to the gold graphs or have been aligned at the node level.

The Smatch algorithm used for evaluation of AMR parsers is a randomly initialized, greedy, hill-climbing algorithm. Due to the greedy nature of the algorithm, multiple random restarts are needed to get a stable matching score. Moreover, the problem itself is NP-complete and even its greedy implementation becomes prohibitively slow as the size of graph increases. As a result, Smatch in its original form is quite inefficient for document-level AMR evaluation.

\subsection{Smatch Evaluation}\label{sec:smatch}

Smatch \citep{smatch} views an AMR graph as a set of triples, where each triple comprises of either a pair of nodes with a relation label (i.e.~edges in the graph) or one node with an attribute and its value (i.e.~concepts and attributes). Given the set of triples for a source and a target graph and an \textit{optimal} alignment between their nodes, the metric computes the precision, recall and F-score over triples between the two graphs. An \textit{optimal} alignment is the one that will maximize the F-score. \Citet{smatch} provide a greedy implementation of Smatch that uses a combination of a `smart initialization' and greedy hill-climbing steps to get closer to an optimal mapping. Each greedy step needs to evaluate $m(n-1)$ options where $m$ and $n$ are the numbers of nodes in the two graphs being matched. Therefore, Using this implementation of Smatch at the document-level is very slow. 


\subsection{Document-level Smatch}

The sentence-level Smatch implementation\footnote{\url{https://amr.isi.edu/evaluation.html}} relies on a pool of candidate node-mappings to efficiently select the next greedy step. The pool includes all the node mappings that can possibly add to the triple F-score.
We can restrict this candidate pool further for document-level Smatch if the source and target documents are aligned at the sentence level. More specifically, we can forbid any source to target node alignments where the sentence(s) of the source node is(are) not aligned with the sentence(s) of the target node. Note, that in \repname graphs, certain nodes can belong to multiple sentences such as \amr{coref-entity} or a merged node. 


We propose an implementation of Smatch (\evlname) that assumes alignment between the roots of sentence-level subgraphs of a pair of document-level AMR graphs. Nodes are first categorised by sentences with each node possibly assigned to multiple sentences. Next, the candidate mappings pool is constructed respecting the sentence-level alignments. In particular, a node in one AMR cannot be mapped to a node in the other AMR if none of their assigned sentences are aligned (see \cref{algo} for details).
For instance, consider the example in \cref{fig5}(c) -- the node for the concept \textbf{\amr{now}} must be aligned to a node in the third sentence in a target graph, whereas the \textbf{\amr{person}} node, merged from mentions in all 3 original sentences, is not constrained\footnote{Note that there could be a case where the predicted parse of sentence~2 resembles the correct parse of sentence~1 and vice versa without any coreference link between the two---then the proposed constraint would prevent finding the optimal alignment. However, this happens quite infrequently. Moreover, we argue that accidental mapping of triples between graphs of entirely unrelated sentences should not be rewarded. Note also that for the border case where all nodes are connected with all sentences, the constrained version will be same as the original Smatch. 
}.

\evlname allows us to evaluate the \repname development set comprising of 42 documents in roughly 4 minutes with the default four random-restarts. This is a manageable time frame for the purpose of parser comparisons. The original Smatch evaluation for the same setup ran out-of-memory (with up-to 200GB allocation) without a result. \Cref{tab:docsmatch} compares the Smatch scores and the runtimes of our implementation with those of the original Smatch with 1 random restart. 


\subsection{Coreference Subscore}\label{sec:subscore}

A side effect of representing and evaluating the document AMR as a single unified graph is that we can not analyze the coreference performance of the parser separately. To mitigate this, we propose and implement a breakdown of Smatch that provides a separate coreference subscore. For the purpose of coreference subscore, all nodes connected to multiple sentences are considered coreferent nodes. Incoming edges for each coreferent node are counted as a part of the coreference subscore, as well as bridging relations and nodes with the labels \amr{coref-entity} or \amr{interloculor-entity}. Note, in the case of merged nodes, their incoming edges count towards the coreference scores, however the node themselves (i.e.~their instance triples) are not counted towards the coreference score.

\begin{table*}[t]
\setlength{\tabcolsep}{5.5pt}
\begin{center}
\begin{tabular}{@{}cc|ccc|ccc|ccc@{}}
\toprule 
\multicolumn{2}{c|}{MS-AMR Splits} & 
\multicolumn{3}{c|}{Double Anno. (Dev1)} &
\multicolumn{3}{c|}{Double Anno. (Dev2)} &
\multicolumn{3}{c}{Test}
\\
\midrule 
AMR    & Coref    & Smatch & Coref & Reent & Smatch & Coref & Reent & Smatch & Coref & Reent \\
\midrule 
Gold   & None     & 87.6 & 0  & 72 & 88.6 & 0  & 73 & 86.4  & 0  & 72 \\
Gold   & CoreNLP  & 89.7 & 34 & 76 & 90.6 & 35 & 78 & 90.6  & 47 & 80 \\
Gold   & AllenNLP & 90.5 & 40 & 78 & 91.0 & 41 & 79 & 91.3  & 53 & 82 \\
\midrule
\midrule
\multicolumn{2}{l|}{\cite{anikina2020crac}} & - & - & - & - & - & - & 44.3        & 17   & 21 \\  
\midrule
S-BART & None     & 67.1 & 0  & 53 & 67.8 & 0  & 53 & 67.5 & 0  & 55 \\
S-BART & CoreNLP  & 68.7 & 28 & 57 & 69.3 & 29 & 57 & 71.3 & 43 & 63 \\
S-BART & AllenNLP & \textbf{69.4} & \textbf{33} & \textbf{59} & \textbf{69.8} & \textbf{34} & \textbf{59} & \textbf{72.0} & \textbf{50} &  \textbf{65} \\

\bottomrule
\end{tabular}
\end{center}
\caption{Document-level Smatch, coreference sub-scores (Coref) and reentrancy scores (Reent) on MS-AMR double annotations (Dev1 and Dev2) and test splits -- using various combinations of gold and predicted AMR graphs with predicted coreferences from CoreNLP \cite{clark2016improving} and AllenNLP \citep{joshi2020spanbert}. 
}
    \label{tab:results}
\end{table*}

Since what is considered a coreference depends on the graph structure, the edges and nodes that are considered coreferent in the gold graph might be different from those in the predicted graph. Therefore, to calculate F1 for the coreference subscore, we consider an edge or node to be a correct match if (1) it has a matching node or edge according to the standard Smatch score, and (2) the node or edge is part of coreference in both the gold and predicted graph. Recall is calculated as a percentage of gold coreference nodes/edges, precision is a percentage of predicted coreference nodes/edges, and F1 is taken as the harmonic average.





\section{Experiments and Results}\label{sec:parsing}

\begin{table}[t]
\begin{center}
\begin{tabular}{l|c|cc|cc}
\toprule 
\multicolumn{2}{c|}{Impl.} & 
\multicolumn{2}{c|}{Original } & 
\multicolumn{2}{c}{\evlname} \\
\midrule
Split & R & Time & Smatch & Time & Smatch \\
\midrule
Dev1 & 4 & - & - & 244 & 69.4 \\ 
Dev2 & 4 & - & - & 136 & 69.8 \\ 
Test & 4 & - & - & 417 & 72.0 \\
\midrule
Dev1 & 1 & \hphantom{0}927  & 69.5 & \hphantom{0}66 & 69.3 \\ 
Dev2 & 1 & \hphantom{0}945  & 69.8 & \hphantom{0}41 & 69.7 \\ 
Test & 1 & 1314 & 71.3 & 104 & 72.0 \\

\bottomrule
\end{tabular}
\end{center}
\caption{Comparison of Smatch scores and runtimes (in seconds) between the original Smatch implementation (Original) and our proposed implementation  (\evlname). All results are on our best performing pipeline system. R is the number of random restarts. `Original' Smatch runs out of memory for R$>$1. 
}
    \label{tab:docsmatch}
\end{table}

We use \repname along with our efficient implementation of Smatch to assess the quality of two document-level AMR parsing systems. First, we develop a pipeline system combining a top-performing AMR parser \citep{zhou2021structure} and a state-of-the-art coreference resolution system \citep{joshi2020spanbert}. We also provide the pipeline results with CoreNLP's neural coreference resolution system \cite{clark2016improving} v4.3.2 for additional point of comparison. Second,  we reevaluate the past best system of \citet{anikina2020crac} -- this is also a pipeline approach combining AMR parser of \citep{lindemann-etal-2019-compositional} with AllenNLP coreference resolution system\footnote{We obtained the document-level graphs before merge operations from \citet{anikina2020crac} for the purpose of re-evaluation in \repname format.}.

\subsection{Our Pipeline Approach}


We use the BART-based structured transformer model of \citet{zhou2021structure} to produce sentence-level AMR graphs. In particular, we use the StructBART-S version of their system referred to as S-BART in \cref{tab:results}. This parser produces node-to-token alignments as part of its output -- we use these alignments to match coreference systems' outputs with AMR graphs. Text-based coreferences are obtained using the systems of \citet{joshi2020spanbert} and \citet{clark2016improving}. Coreference chains are computed from these prediction files.\footnote{Using the package \url{https://github.com/boberle/corefconversion}} 

In order to incorporate this coreference information into the AMR graphs, we first convert node-to-token alignments into node-to-span alignments. The span of a node is defined as the smallest text span containing all the tokens aligned to any of its descendants (to avoid loops, re-entrant edges are removed keeping only the first). With node-to-span alignments, a predicted mention is assigned to the node with the shortest span containing the mention. If there is more than one candidate node, the one with the greatest height, subsuming the other candidates is selected.
Instances of coreference within a sentence are ignored assuming that the sentence-level parser has already taken care of them.

\subsection{Results}

\Cref{tab:results} shows the results on MS-AMR double annotation documents (used here as development set) and its test split. Both gold and predicted AMR graphs are converted to \repname before running Smatch evaluation. All numbers are produced using document-level Smatch with 4 random restarts. 

Our pipeline approach outperforms the previous best system by a large margin, providing a strong baseline for future research on this task. This is due mainly to difference in quality of the underlying sentence-level parsers. 

Note that significant improvements in coreference quality result in only small improvements in overall Smatch score -- showing that a separate coreference subscore is essential for assessing a system's performance on cross-sentential relations. We also report reentrancy scores \cite{damonte2016incremental} for comparison. While the performance gap is more pronounced in reentrancy scores compared to overall Smatch, the gains are best highlighted in coreference subscore. For instance, reentrancy scores improves by up to 2 points from CoreNLP to AllenNLP in all settings -- coref subscores, on the other hand, shows up to 7 points improvement giving a finer range for coreference evaluation. 



\paragraph{Impact of Representation on Smatch:} To highlight how document-level  representation can affect Smatch scores and efficiency, we compare the gold double-annotation development set with and without coreference links. In addition to \repname we consider three representations: 1) No-Merge: where all coreference nodes are linked via coref-entity without any merging 2) Merge-NE: where only Named Entities are merged and 3) Merge-All: where all coreference nodes in a chain are merged \cite{ogorman-etal-2018-amr}. One of our aims for \repname was to ensure that the lack of coreference links is visible in the overall Smatch score. \Cref{tab:docsmatch2} shows that \repname makes this gap bigger without losing efficiency or semantic information. 

\begin{table}[t]
\begin{center}
\begin{tabular}{l|cc}
\toprule 
Dev1  & {Smatch} & {Time (s)}  \\
\midrule
No-Merge  & 93.3 & \hphantom{0}66 \\ 
Merge-NE  & 92.8 & \hphantom{0}66 \\
\repname  & 87.6 & \hphantom{0}90 \\
Merge-All & 82.4 & 153 \\
\bottomrule
\end{tabular}
\end{center}
\caption{Impact of representation on evaluation scores and runtimes. Comparing double-annotation gold documents (Dev1) with and without coreference links in different document-level representations.
}
    \label{tab:docsmatch2}
\end{table}




\section{Related Work}

\textbf{MS-AMR Annotations} MS-AMR annotations by \citet{ogorman-etal-2018-amr} include coreference chains, implicit roles and bridging relations.
In the context of AMR-based summarization, \citet{lee2021amrsummarization} present a novel dataset consisting of human-annotated alignments between the nodes of paired documents and summaries to evaluate merge strategies for merging individual AMR graphs into a document graphs. However, they sought out the merge operations that can serve as cross sentential coreference in the absence of any annotations -- they only merge nodes with same surface forms, except for 'person' nodes. Our work, on the other hand, outlines a representation for already available gold annotations, where nodes' surface forms don't match in a large number of cases.


\textbf{MS-AMR Evaluations and Models} \Citet{ogorman-etal-2018-amr} proposes Smatch as primary method for scoring  MS-AMRs. They also report CoNLL-F1 relying on Smatch alignments. Adopting the methods from \citet{ogorman-etal-2018-amr}, \citet{anikina2020crac} presented a comparative evaluation of various coreference resolution systems over MS-AMR test sets and document-level Smatch evaluations of machine generated sentence-level AMRs augmented with coreference predictions from various systems.
The best approach from their study is incorporated as a baseline in \cref{sec:parsing}. \Citet{fu2021amrcoref} introduce an AMR coreference resolution system that uses graph neural network to model gold sentence-level AMR graphs for coreference predictions. This system assumes gold graphs and is not comparable with document-level parsing systems. Use of gold graphs also alleviates the need for alignments between gold and predicted graphs for the purpose of evaluation. \Citet{bai2021amrdialog} constructed dialogue-level AMR graphs from multiple utterance level AMRs by incorporating inter-sentence coreference, speaker and identical concept information into sentence-level AMRs.



\section{Conclusion}

We have presented \repname, a graph representation for document-level AMR graphs based on coreference annotations. 
Relative to the original sentence-level graphs, \repname removes redundancy 
without information loss.
We modified the implementation of Smatch to take advantage of sentence provenance 
to efficiently search for node alignments when comparing two document-level graphs. 
Finally, we reported results for a document-level parsing pipeline that can serve as a strong baseline for future work on this task. 

\section*{Acknowledgment}

Jeffrey Flanigan was supported by the NSF National AI Institute for Student-AI Teaming (iSAT) under grant DRL 2019805. The opinions expressed are those of the authors and do not represent views of the NSF.


\bibliography{custom}
\bibliographystyle{acl_natbib}

\appendix
\section{Constrained Candidate Node Mappings for \evlname}
\label{algo}

\begin{algorithm}\small
  \caption{Constrained candidate node mappings for efficient document-level Smatch computation.}
  \textbf{inputs:} \\
  $dAMR_1, dAMR_2$ pair of AMRs for a document \\
  $sRoots_1, sRoots_2$ aligned sentence roots
  \hrule
  \begin{algorithmic}
  \STATE $N\gets \textit{Number of sentences}$ \COMMENT{in the document}
  \COMMENT{collect descendant of $sRoots$}
  \FOR{$i \gets 1..N$} 
   \STATE $Desc_1[i]\gets \textsc{getDesc}(dAMR_1,sRoots_1[i])$
   \STATE $Desc_2[i]\gets \textsc{getDesc}(dAMR_2,sRoots_2[i])$
  \ENDFOR
    \STATE $CandMap \gets \{\}$ \COMMENT{Candidate Node Mappings}
    \FOR{$i \gets 1..N$}
        \FOR{$node\in Desc_1[i]$}
            \STATE $CandMap[node] \mathrel{{+}{=}} Desc_2[i]$
        \ENDFOR
    \ENDFOR
    \STATE \textbf{return} $CandMap$
  \end{algorithmic}
\end{algorithm}







\end{document}